\documentclass[a4paper,twoside]{article}

\usepackage{epsfig}
\usepackage{subfigure}
\usepackage{calc}
\usepackage{amssymb}
\usepackage{amstext}
\usepackage{amsmath}
\usepackage{amsthm}
\usepackage{multicol}
\usepackage{pslatex}
\usepackage{apalike}

\usepackage{times}
\usepackage{latexsym}
\usepackage{graphicx}
\usepackage{url}
\usepackage{algorithm}
\usepackage{algpseudocode}

\usepackage{SCITEPRESS}     

\begin{document}

\title{Detecting and assessing contextual change in diachronic text documents using context volatility }

\author{\authorname{Christian Kahmann\sup{1},Andreas Niekler\sup{1} and Gerhard Heyer\sup{1}}
\affiliation{\sup{1}Leipzig University / Augustusplatz 10, 04109 Leipzig}
\email{\{kahmann,aniekler,heyer\}@informatik.uni-leipzig.de}
}

\keywords{Context Volatility, Semantic Change}

\abstract{
Terms in diachronic text corpora may exhibit a high degree of semantic dynamics that is only partially captured by the common notion of semantic change.
The new measure of \textit{context volatility} that we propose models the degree by which terms change context in a text collection over time.
The computation of context volatility for a word relies on the significance-values of its co-occurrent terms and the corresponding co-occurrence ranks in sequential time spans.
We define a baseline and present an efficient computational approach in order to overcome problems related to computational issues in the data structure.
Results are evaluated both, on synthetic documents that are used to simulate contextual changes, and a real example based on British newspaper texts. 
\textbf{The data and software are avaiable at https://git.informatik.uni-leipzig.de/mam10cip/KDIR.git}
}

\onecolumn \maketitle \normalsize \vfill

\section{\uppercase{Introduction}}

When dealing with diachronic text corpora, we frequently encounter terms that for a certain span of time exhibit a change of linguistic context, and thus – in the paradigm of distributional semantics – exhibit a change of meaning.
For applications in information retrieval and machine learning tasks this causes problems because terms then are not unambiguous, hindering the task of retrieving, or structuring, relevant documents or information.
However, understanding semantic change can also be a research goal on its own, such as work in historical semantics \cite{simpson_oxford_1989}, or in the digital humanities where semantic change has been used as a clue to better understand political, scientific, and technical changes, or cultural evolution in general \cite{michel_quantitative_2011,wijaya_understanding_2011}.
In information retrieval, finding terms that significantly change their meaning over some period of time can also be a key to exploratory search \cite{heyer_interaktive_2011}.

While there is plenty of work on how to detect and describe semantic change of particular words, for example ``gay'', ``awful'', or ``broadcast'' in English between 1850 and 2000 \cite[cf.]{hamilton_diachronic_2016} by highlighting the differences in context as derived by co-occurrence analysis \cite[cf.]{jurish_diacollo:_2015} or topic models \cite[e.g.]{jahnichen_exploratory_2015}, it is still an open question of how to identify those terms in a diachronic collection of text that undergo – by some degree – a change of context, and thus exhibit a semantic change in the paradigm of distributional semantics.
In what follows, we present context volatility as a new and innovative measure that captures a term's rate of contextual change during a certain period of time. 
The proposed measure allows to specify the degree of a term's contextual changes in a document collection over some period of time, irrespectively of the amount of text.
This way we are able to identify terms in a diachronic corpus of text that are semantically stable, i.e. that undergo little or no changes in context, as well as terms that are semantically volatile, i.e. that undergo continuous or rapid changes in their linguistic context.
Often, semantically volatile terms are highly controversial, such as ``Brexit`` in the 2016 British public debate.
A term's context volatility complements its frequency, a feature that is of particular interest when we are interested in detecting weak signals, or early warnings, in the temporal development of a corpus related to low frequent terms indicating subsequent semantic change.
Context volatility is related to the notion of volatility in financial mathematics \cite{taylor_introduction_2007}, and we can draw a rough analogy that just like the rate of change in the price of a stock is an indication of risk, a high degree of context volatility of a term is an indication that the fair meaning of a term is still being negotiated by the linguistic community. 

We begin in section 2 by showing how our approach differs from other procedures in the spirit of distributional semantics.
We then explore and evaluate different forms of contextual change and identify computational problems that arise when applying context volatility to diachronic text corpora.
By evaluating the basic intuition and strategies to overcome the difficulties we shall then present in section 3 an approach which robustly calculates context volatility without any statistical bias, and in section 4 introduce a novel algorithm which avoids sparsity problems on diachronic corpora, the MinMax-algorithm.
To evaluate the measure, we apply the algorithm to a synthetic data set in section 5, based on a distinction between three cases of how the context of a word may change, viz. (1) a change in the probability of co-occurring terms, (2) the appearance of new contexts, and (3) the disappearance of previous contexts.
Finally we apply our measure to real life data from the Guardian API and illustrate its usefulness in contrast to a purely frequency based approach with respect to the term ``Brexit''.

\section{\uppercase{Related Work}}\label{sec:related}

Several studies address the analysis of variation in context of terms in order to detect semantic change and the evolution of terms.
Three different areas to model contextual variations can be distinguished: (1) methods based on the analysis of patterns and linguistic clues to explain term variations, (2) methods that explore the latent semantic space of single words, and (3) methods for the analysis of topic membership.
\cite{jatowt_framework_2014} use the latent semantics of words in order to create representations of a term's evolution which is a similar information as used in context volatility.
%
%
The approach models semantic change over time by setting a certain time period as reference point and comparing a latent semantic space to that reference over time.
\cite{fernandez-silva_picturing_2011,mitra2014} or \cite{picton_proposed_2011} look for linguistic clues and different patterns of variation to better understand the dynamics of terms. 
%
%
The popular word2vec model has also been utilized for tracking changes in vocabulary contexts \cite{KimCHHP14,Kenter15}. 
Both word2vec-based approaches use references in time or seed words to emphasize the change but do not quantify it independently to the reference.
\cite{KimCHHP14} add the quantity of the changes throughout growing time windows towards a global context representation but do not examine the possibility of detecting different phases in the intensity of change whereas \cite{Kenter15} produce changing lists of ranked context words w.r.t. the seed words.
One major concern in suing the word2vec approaches is the fact that those models require large amounts of text.
Both described approaches fulfill this requirement by a workaround to artificially boost the amount of data.
This influences the ability to quantify and describe contextual changes according to the observable data.
Assuming a Bayesian approach, topic modeling is another method to analyze the usage of terms and their embeddedness within topics over time \cite{blei_dynamic_2006,zhang_evolutionary_2010,rohrdantz_towards_2011,rohrdantz_lexical_2012,jahnichen_time_2016}.
%
%
%
%
%
However, topic model based approaches always require an interpretation of the topics and their context.
In effect, the analysis of a term's change is relative to the interpretation of the global topic cluster, and strongly depends on it.
%
%
In order to identify contextual variations, we also need to look at the key terms that drive the changes at the micro level.
Context volatility differs from previous work in its purpose because it does not start with a fixed set of terms to study and trace their evolution, but rather detects terms in a collection of documents that may be indicative of contextual change for some time.
The notion of context volatility is introduced in \cite{heyer_change_2009,hutchison_towards_2010,heyer_modeling_2016}.
The respective works present single case studies to evaluate the plausibility of the proposed measure.
Several negative effects in the data are not discussed and evaluated. 
For example, the appearance of new contexts or the absence of certain word associations in single time slices cause gaps in the co-occurrence rank statistics which were ignored in those works.
%
%
%
The measure determines the quantity of contextual change by observing the coefficient of variance in the ranks for all word co-occurrences throughout time.
Additionally, it does not use latent representations but the co-occurrence information itself without any smoothing.
The information of temporarily non-observable co-occurrences is present and can be used for the determination of contextual change directly.
In section~\ref{sec:cv} we take up on those effects that directly influence the procedure and define alternative approaches to overcome associated problems.
In sum, while related work on the dynamics of terms usually starts with a reference (like pre-selected terms, reference points in time, some pre-defined latent semantics structures, or given topic structures), context volatility aims at automatically identifying terms that exhibit a high degree of contextual variation in a diachronic corpus regardless of some external reference or starting point.
The measure of context volatility is intended to support exploratory search for central terms\footnote{The notion of centrality of terms is used in \cite{picton_proposed_2011}. It captures the observation that central terms simultaneously appear or disappear in a corpus when the key assumptions, or consensus, amongst the stakeholders of a domain change} in diachronic corpora, in particular, if we want to identify periods of time that are characterized by substantial semantic transformation.
However, we do not claim that the measure quantifies meaning change or semantic change, the measure quantifies the dynamics of a term's contextual information within a diachronic corpus.

\section{\uppercase{Definition of Context Volatility}}\label{sec:cv}

The computation of context volatility is based on term-term matrices for every time slice derived from a diachronic corpus \cite{heyer_modeling_2016}.
Those matrices hold the co-occurrence information for each word $w$ in a time slice $t$.
One can use different significance weights based on the co-occurrence counts such as the log-likelihood-ratio, the dice measure or the mutual information to represent the co-occurrences \cite{bordag_comparison_2008}.
%
%
First, the corpus is divided into sets of documents belonging to equal years, months, weeks, days or even hours and minutes to define the time slices $t$.
The set of all time slices is $T$.
It is necessary to determine for every word $w$ of the vocabulary $V$ and every time slice $t$ the co-occurrences $\mathbf{w*}$, e.g. a term-term matrix $C_t$ with the co-occurrence counts.
The matrix has dimension $V \times V$.
Based on the counts in $C_t$ a significance-weight can be calculated on all word pairs which results in a term-term matrix of significant co-occurrences $S_t$.
The significance-values for the co-occurrences of a word $w$ in $t$ are sorted to assign a ranking to the co-occurrences of a word. 
For every word which co-occurs with $w$ in $t$ this ranking may differ in the time slices or the ranking can't be applied if this very co-occurrence is not observable. 
The ranks are stored in a matrix $R_t$.
Based on consecutive time stamps the ranks for every word pair in all $R_{w,\mathbf{w*},t}$ build a sequence.
The transition of a word pairs ranking through the time slices is quantified by the coefficient of variance on the ranks.
It is possible to perform this calculation only on a history $h$ of time slices which gives the variance in the ranking of a word pair for a shorter time span.
The context volatility for a word $w$ in $t$ is then the mean of all rank variances from the word pairs $w,\textbf{w*}$.
Note, that not all combinations in $w,\textbf{w*}$ have a count $> 0$ in all time slices simply because not every word co-occurs with every other word all the time.
There are words which never occur together with $w$ or just in some time slices.
We therefor define the basic measure of context volatility of a term as an averaged operation $cv(C_{w,i},h)$ on all co-occurences of $w$ where $C_{w,i}$ is the i\textsuperscript{th} co-occurrence of $w$.
In the basic setting the mean is build over all co-occurrences $\textbf{w*}$ of $w$ which can be observed in at last 1 time stamp in $h$.
Precisely, we can define the final calculation of the volatility for $h$ as
\begin{equation}
CV_{w,h}=\frac{1}{\|C_{w,h}\|}\sum_{i}cv(C_{w,i},h),
\label{eq:}
\end{equation}
with $\|C_{w,h}\|$ the number of co-occurrences on $w$ which could be observed in $h$.
For consecutive histories the context volatility of $w$ forms a time series where each data point contains the context volatility at a time slice $t$ for a given history $h$.
This represents the mean context volatility for all contextual information about a word and we get an average change measure for the co-occurrences.
Informally, context volatility computes a term's change of context by averaging the changes in its co-occurrences for a defined number of time slices.
Alternatively, $h$ could be set to all time slices in $T$ to produce a global context volatility for the words in a diachronic text source.

\subsection{{Limitations}}\label{sec:lim}
So far the main limitation of context volatility has been an adequate handling of gaps.
When applying context volatility as described in \cite{hutchison_towards_2010} it can be shown that there are many cases for which a co-occurrence of 2 words at a specific point of time not only changes in usage frequency.
%
%
The effect could be observed in cases for which new vocabulary is associated with a given word $w$ in diachronic corpora or some contexts are temporarily not used.
This does not necessarily mean that the absence of a context  introduces a lasting change to the semantic meaning of a word.
But both cases contain important information about the dynamics in the contextual embedding of a word.
Thus, the resulting gaps in the rank sequence of diachronic co-occurrences for a word must be handled accordingly to prevent a bias.
Different strategies for the handling of those gaps seem plausible.
For example, consider the situation where we calculate the variance of 10 consecutive ranks, e.g. $h$ is set to 10 time slices, of a co-occurring word which is given by $R_{w,i,h} = {1,X,2,X,3,X,4,X,5,X}$. 
The $X$ represents a gap, e.g. a co-occurrence count of 0 in the according time slice.
If the volatility of this progression should be calculated as in \cite{hutchison_towards_2010} we would calculate the coefficient of variation of all ranks which are observable.
In the experiments the authors used a very large corpus with a large $h$ and the influence of the gaps is presumed to produce a small bias.
However, for smaller corpora or smaller amounts of documents for a time slice we can't ignore the influence of such decisions.
Often, a co-occurrence can only be observed in a minority of the time slices if, for example, the corpus is reduced to a set of documents containing a specific topic.
Following the definition of \cite{hutchison_towards_2010} we set 
\begin{equation}
	cv(C_{w,i},h) = \frac{\sigma(R_{w,i,h})}{\bar R_{w,i,h}}.
\label{eq:iqr}
\end{equation}
Likewise, we can use the significance-values for co-occurring terms directly and set
\begin{equation}
cv(C_{w,i},h) = \sigma(S_{w,i,h}),
\label{eq:sig}
\end{equation}
where $\sigma(S_{w,i,h})$ is the standard deviation of the i\textsuperscript{th} co-occurrence significances of $w$ in the history $h$. 
We use the standard deviation since the significance-values are not linearly distributed and a major amount of co-occurrences has very little significance-values.
Using the coefficient of variance would produce large values in changes of small significances which is an undesired behavior.

To carry out this calculations in a similar manner like \cite{hutchison_towards_2010} only values of $R_{w,i,h}$  are included which can be observed in the data, e.g. $R_{w,i,h} = {1,2,3,4,5}$.
%
We do not include the non-observable co-occurrences with $0$ or the maximum possible rank, e.g. the count of all possible co-occurrences or the vocabulary, to fill the gaps because this introduces an undesired bias.
A better strategy to handle the non-observable ranks is to set the missing co-occurrence in $t$ by other information.
We calculate all co-occurrence significances on all documents on all time slices first, in order to have a ``global`` co-occurrence statistic $S^G$.
If a significance value in a time stamp $S_{w,i,t}$ cannot be observed we set this value to $S_{w,i}^G$ if this significance is greater than 0, e.g. the co-occurrence can be observed in some time stamp in $T$ but not in all.
This deletes the gaps and introduces a global knowledge about the contexts. 
Co-occurrences which are new or emerging or just observable in some time stamps are somehow of higher significance in the time stamp but of lower significance w.r.t. the whole corpus.
This procedure prevents a bias and numerical problems with missing ranks or significances.
However, the dynamics in the co-occurrence statistics are still determinable.
In section~\ref{sec:eval} we evaluate the baseline method by using the ranks and measure their coefficient of variance by ignoring missing information (Baseline).
Additionally, we test the same setup using the significances directly (Sig).
Both setups are evaluated once again with beforehand added global co-occurrence information(GlobalBaseline, GlobalSig).
\section{\uppercase{MinMax-Algorithm}}\label{sec:minmax}
In contrast to the notion of context volatility, where the gaps were either not used at all or replaced by global information, the MinMax-algorithm does use the gap information itself.
The difference $d$ in the ranks of a co-occurrence $w,\textbf{w*}$ is measured from time slice to time slice separately, then summed up and divided by the number of time slices considered.
This results in a mean distance between the ranks of the time slices w.r.t.  $w$ and $h$.
A major distinction is the introduction of 2 different ranking functions (formula \ref{rank_not_zero} and \ref{rank_zero}) when setting the ranks for all co-occurences $w, \mathbf{w*}$ in a time slice $t$.
We apply both formulas to all observable co-occurrences resulting in 2 ranks per co-occurrence.
Formula~\ref{rank_not_zero} applies the ranks decreasing from the maximum number of co-occurrences $w$ has in a time slice considered in $h$.
Words $\mathbf{w*}$ with significance-values of 0, e.g. gaps, share rank 0. 
In formula~\ref{rank_zero} the ranks are assigned decreasing from the number of co-occurrences $w$ has in time slice $t$.
Significances with value 0 again share rank 0.
In the following equations $R_{w,\mathbf{w*},t}$ is the list of co-occurrence ranks for word $w$ w.r.t. time slice $t$ and $R_{w,i,t}$ the rank of the i\textsuperscript{th} co-occurrence for $w$ w.r.t. $t$.
The quantity $max(R_{w,\mathbf{w*},1 \ldots h})$ is the maximum rank an observable co-occurrence of $w$ can take in a history $h$, e.g. the maximum number of co-occurrences of $w$ amongst the time slices included in $h$.
Additionally, this quantity is utilized to normalize the determined ranks in the interval [0,1].
The normalization removes the strong dependence towards a high number of co-occurrences because the higher the number of co-occurrences, the more likely it is to see a higher absolute change for the ranks.
This enables the comparison of words with big differences in their frequencies and with that in their number of co-occurrences.
\begin{equation}
R^{\neg0}(R_{w,\mathbf{w*},t})=\frac{max(R_{w,\mathbf{w*},1 \ldots h})+1-R_{w,i,t}}{max(R_{w,\mathbf{w*},1 \ldots h})}
 \label{rank_not_zero}
\end{equation}
%
%
\begin{equation}
R^{0}(R_{w,\mathbf{w*},t})=\frac{max(R_{w,\mathbf{w*},t})+1-R_{w,i,t}}{max(R_{w,\mathbf{w*},1 \ldots h})}
\label{rank_zero}
\end{equation} 
Consequently, besides having an absolute maximum (rank for highest significance) we now confirm having an absolute minimum (rank 0) over all time slices as well.
The gap-information is used directly because when new words appear, we are able to measure the distance between rank 0 and the relative rank a new appearing word has in $t+1$.
%
%
%
The ranks of formula~\ref{rank_not_zero} are used when neither of the 2 consecutive entries to calculate $d$ represent a gap ($R^{\neg 0}$).
When either one or both regarded entries represent a gap we use the ranks determinded by formula~\ref{rank_zero} ($R^{ 0}$).
We can summarize the procedure of calculating the mean distance of all  $w,\textbf{w*}$ concurrent words to
\begin{equation}   CV_{w,h}=\frac{1}{\|C_{w,h}\|\cdot (h-1)} \sum_{i=1}^{\|C_{w,h}\|} \sum_{t=1}^{h-1} cv(C_{w,i,h}) \label{formula_minmax}    \end{equation}
with $cv(C_{w,i,h})=$
\begin{equation} 
\begin{cases}      \mid R^{0}(R_{w,t,i})-R^{0}(R_{w,t+1,i})\mid & \text{if\ } a   \\ 
				  \mid R^{\neg0}(R_{w,t,i})-R^{\neg0}(R_{w,t+1,i})\mid  & \text{if\ } b   
\end{cases}
\end{equation}
where  condition $a: R_{w,t,i} \lor R_{w,t+1,i}=0 $ and 
	$b:R_{w,t,i} \land R_{w,t+1,i}=\neg0$.

%
\section{\uppercase{Evaluation using a synthetic dataset}} \label{sec:eval}
%
There is no gold standard to validate the quantity of contextual change.
%
Such being the case,  we are utilizing a synthetic data set in which we can manipulate the change of a word's context in a controlled way.
We simulate different situations, where the rate of context change follows clearly defined target functions.
We apply the procedures presented in section~\ref{sec:cv} and~\ref{sec:minmax} to the synthetic data set and compare the results to the target functions we aimed for.
%
%

\subsection{\uppercase{Creating a synthetic data set}}
\begin{figure}[h]
\includegraphics[width=\linewidth]{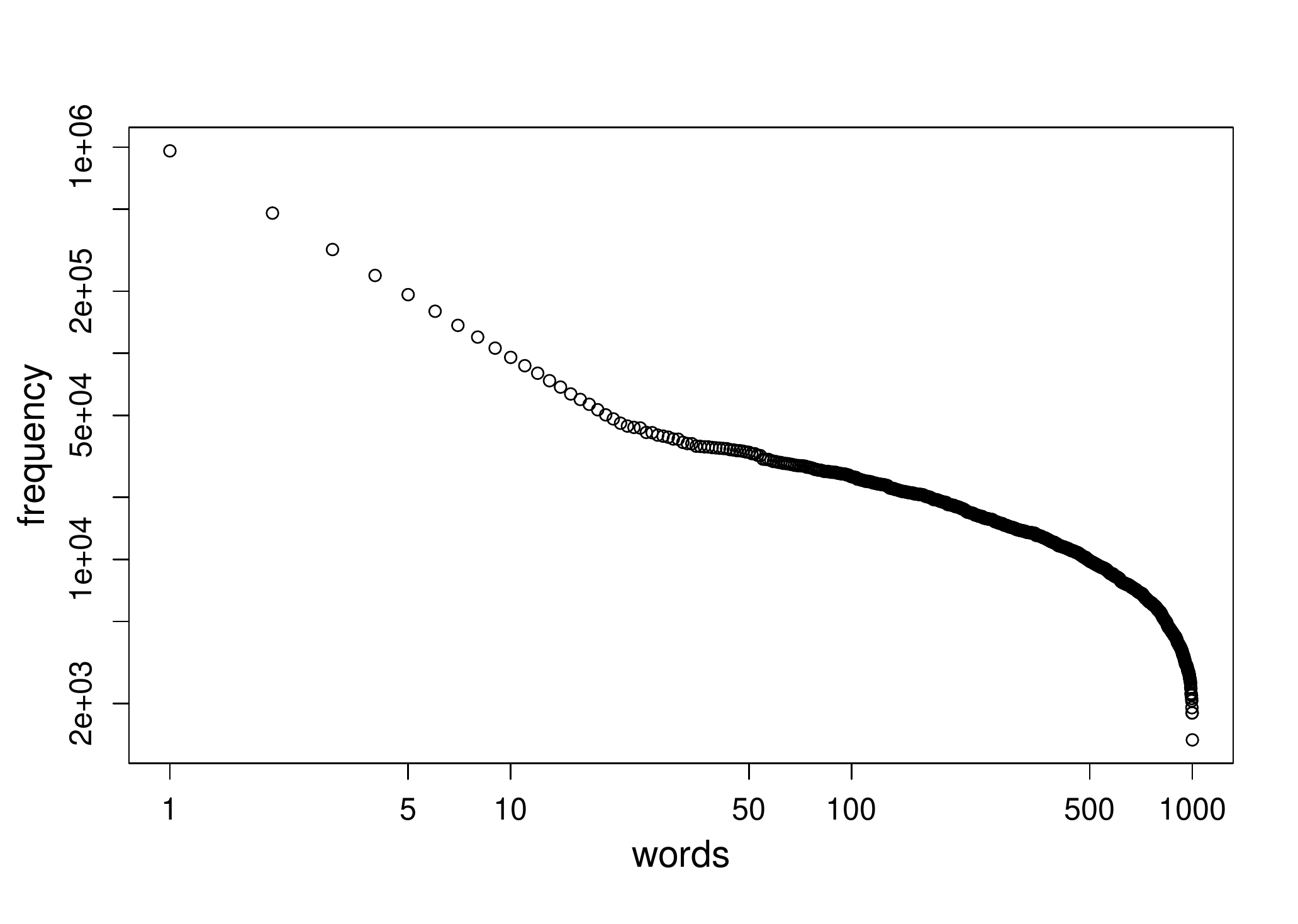}
\caption{Synthetic dataset with double log scale }
\label{zipfdistlog}
\end{figure}

The creation of the data set follows 2 competing goals.
On one hand we want our data set to be as close to an authentic data set as possible.
Therefore, our synthetic data set should be Zipf-distributed in word-frequency which induces noisiness.
On the other hand the context volatility that follows the target functions 
has to be measurable.
Hence, when manipulating the context of a word over time, we can't ensure to not infringe the regulations for a Zipf-distribution.
We create a data set that is somewhat Zipf-distributed but still contains the signals we want to measure (figure \ref{zipfdistlog}) as a trade-off between signal and noise.
The creation of the data set requires a number of time stamps (100), a vocabulary size (1000), the mean-quantity of documents per time stamp (500) and a mean amount of words per document (300).
To simulate the Zipf-distribution, we create a factor $f_{Zipf}^{i}=\frac{1000}{i}$, which assigns to every word $w_{1,...,1000}$ a value indicating a word count.
The factors of counts can be normalized to probability distributions when constructing the documents.
We also define 7 factors $f_{1,\ldots,7}$  responsible for simulating a predefined contextual change.
The values for 7 words $w_{51,\ldots,57}$ are set to 0 in the Zipf-factor ($f_{Zipf}^{51,\ldots,57}=0$) because they'll get boosted in the respective factors $f_{1,\ldots,7}$ and serve as reference on which we evaluate the contextual change. 
Additionally, we designate the first 50 words in the artificial vocabulary, e.g. $w_{1,...,50}$, to act as stopwords.
%
%
In each of the 7 factors we set 150 randomly chosen non-stopwords to simulate an initial context $w,\mathbf{w*}$ and we assign some high values according to $\sim \mathcal{N}(75,25)$.
The values for the reference words $w_{51,\ldots,57}$ in their associated factor is fixed to 200.
All other words in a factor get a value close to 0 (0.1).
The 150 other words ($w,\mathbf{w*}$) in each factor represent the initial words that are very likely to form co-occurrences with their respective fixed word at the first time stamp.
%
The simulation of the contextual change now influences the distribution of the 7 factors $f_{1,\ldots,7}$ from time stamp to time stamp.
Thereby, each factor context is modified following 1 of the target functions triangle ($f_1$), sinus ($f_2$), constant 0 ($f_3$), slide ($f_4$), half circle ($f_5$), hat ($f_6$) and canyon ($f_7$). Illustrations of the functions are  located in the appendix.
Having initiated all factors, the words for every document referring to the first time stamp are sampled.
We use a uniform-distribution to choose 1 factor $j$ among $f_{1,\ldots,7}$ for every document.
Next, we collect a preselected amount of samples for the document using a multinomial distribution 
\begin{equation}p(w_i)=\frac{f_{Zipf}(w_i)+f_{j}(w_i)}{\sum_j^V f_{Zipf}(w_j)+f_{j}(w_j)}\label{multinom}\end{equation} 
over the whole vocabulary.
By adding together the the Zipf-factor and the sampled factor $j$ we are adding artificial noise to the creation of the documents.
%
We could set the influence of the Zipf-factor separately. 
But a higher proportion leads to more noise interfering the context signals.
%
The result is a bag of words constituting each document in the first time stamp.
To proceed to the next time stamp the factors $f_{1,\ldots,7}$ have to be altered to simulate contextual change.
%
%
The target functions control the amount of change in their corresponding factors in dependence to the time stamp to influence the sampling outcome, and consequently the context of the reference words.
Note, the value for the context word in the respective factor stays the same and must stay 0 in all other factors (e.g. $f_1^{w_{51}} = 200,  f_{2,\ldots,7}^{w_{51}} = 0$). 
Besides the definitions of functions to control the amount of contextual change we identify 3 different cases how the context of a word can be influenced.

\begin{enumerate}\item [i.] \textbf{Exchange the probability for co-occurring terms:}
%
%
We model this by taking 2 words from the co-occurrences $w,\mathbf{w*}$ in a factor and swap their probabilities.
This leads to a change of their likelihood to be sampled in the same documents like their context word.

 \item [ii.] \textbf{Appearance of new contexts:}
%
%
To address this, we select 1 word in a factor that has a value close to 0 and replace it with a high value.
As a consequence, when sampling the words, we will likely see a "new" word w.r.t. the context word of a factor.

 \item [iii.]\textbf{Disappearance of contexts:}
%
%
We change the probability value of a co-occurring word (high value) close to 0.
This will likely cause this word to not appear anymore along with a context word.
\end{enumerate} 
In order to create the documents for the following time stamps, the factors $f_{1,\ldots,7}^t$ are influenced in dependence to the the factors of the former time stamp ($t-1$) w.r.t a target function. 
\begin{equation}f_{j}^t\sim change_{targetFunction}(f_{j}^{t-1})\label{factor_change}\end{equation}
%
%
For example if  $f_1$ is following the triangle function, for $t < 50$ the number of exchanged contexts can be determined by the time stamp index $t$.
For time stamps $t > 50$ the number of exchanges is determined $100 - t$.
With the updated factors the succeeding time stamps can be sampled until every document in $T$ is filled.
The final data set for the 7 target functions and all 3 options for contextual change is distributed as shown in figure~\ref{zipfdistlog}.
The data set is not exactly Zipf-distributed because of the forced co-occurrence behavior.
%

\subsection{Results of Evaluation}
\begin{figure}
\includegraphics[width=\linewidth]{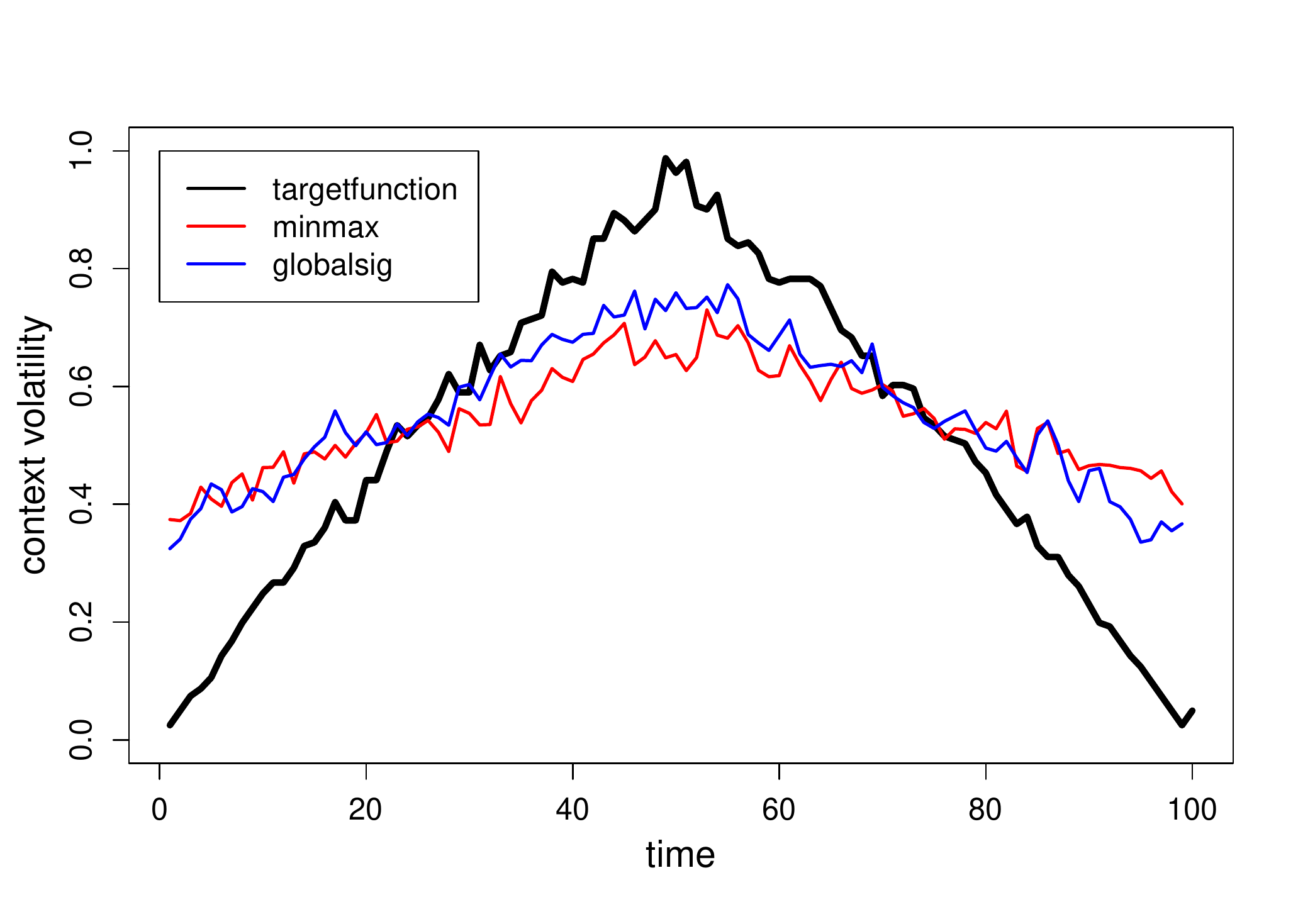}
\caption{Results for MinMax and GlobalSig on first data set }
\label{minmax}
\end{figure}
We created 3 different data sets to test  strengths and weaknesses of the algorithms.
In the first data set (A) all 3 described cases of context change are included (case i., ii., iii.).
The second data set (B) is built by only changing the probability values of already co-occurring terms (case i.).
In the third data set (C) we deploy the options appearance of new co-occurrences (case ii.) and disappearance of known co-occurrences (case iii.).
We use cases (ii.) and (iii.) in the same amount for the data sets A and C, so the quantity of co-occurrences for the respective context words stays constant.
In data set C we refused to add the Zipf-factor, because it interferes the forced co-occurrence gap-effect which we intend to simulate by this configuration.
We applied the context volatility measures to all 3 data sets using a span of $h=2$.
In the interest of a quantifiable comparison we normalized over all results (all 3 data sets) for every context volatility measure.
%
%
We assess the performance of the measures by using the mean distance between the calculated context volatility and the intended target functions of the 7 context words.
An example how the measures approximate the target function can be seen in figure~\ref{minmax}.
\begin{table}[h]
\centering
\resizebox{\linewidth}{!}{%
\begin{tabular}{l|*{6}{c}r}
          & A & B & C & Mean       \\
          \hline
Baseline &0.25&0.11&0.29&0.217\\  
GlobalBaseline &0.35&0.18&0.26&0.263                \\
Sig    & 0.30 & 0.14 & 0.10 & 0.180 \\
GlobalSig & 0.22 & 0.25 & 0.08 & 0.183 \\
MinMax & 0.24 & 0.12 & 0.18 & 0.180 \\
\end{tabular}%
}
\caption{Results using all 3 synthetic data sets; for every method and every data set we calculated the mean distance over all 7 target functions}
\label{results_eval}
\end{table}

\begin{figure}[H]
\includegraphics[width=\linewidth]{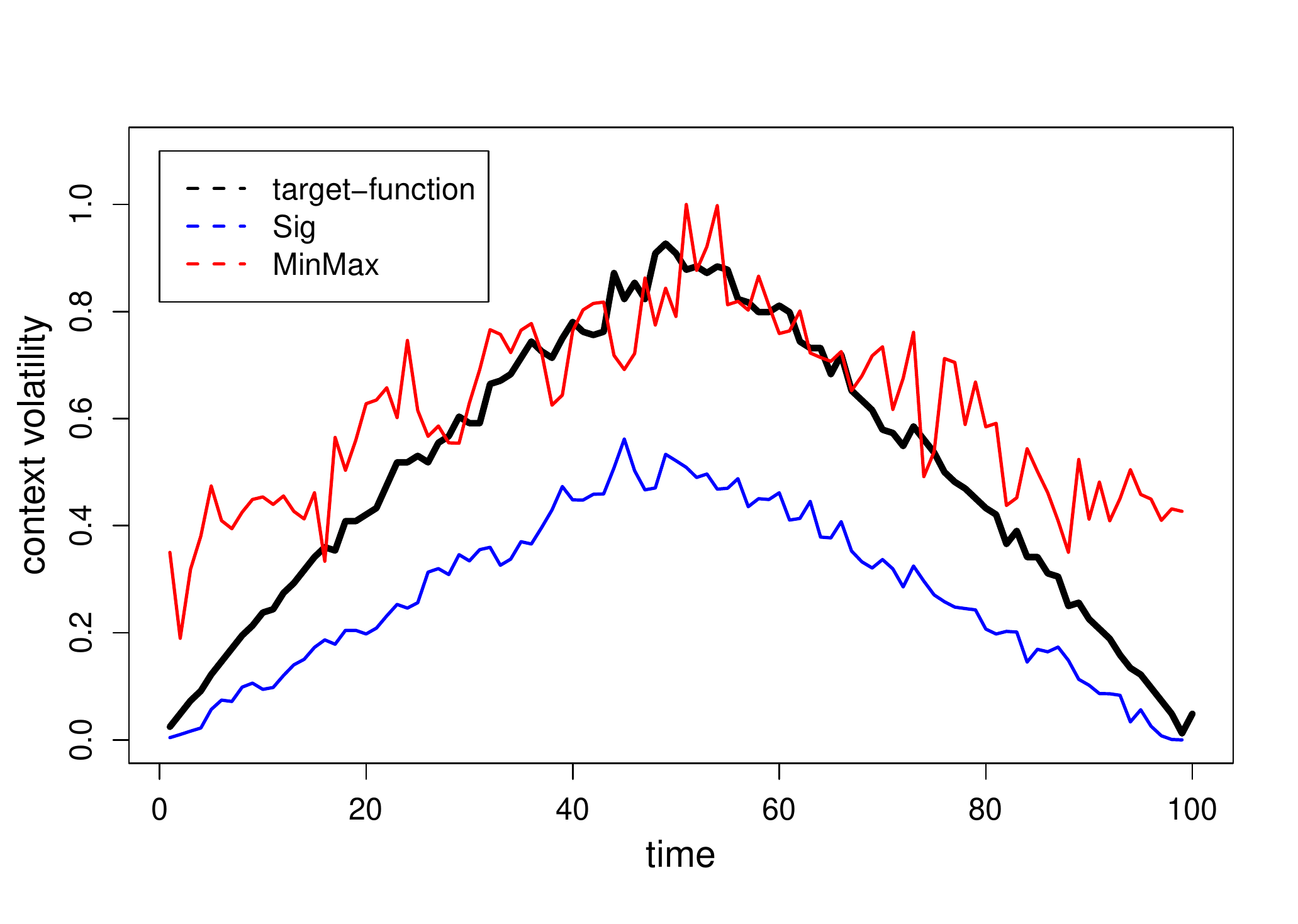}
\caption{Triangle target function and the calculated volatilities using MinMax and Sig with high frequence for reference word}
\label{minmaxsigtriangle}
\end{figure}

All calculation methods are able to detect the function signals for data set A which sends the strongest signals (table~\ref{results_eval}).
The GlobalSig method slightly outperforms all other methods in this setting.
The loss of information that occurs when transferring the significances to ranks might be one cause of the overall worse performance of the baseline methods, and a opportunity to further improve the MinMax method.
When using data set B the MinMax- and the Baseline-algorithm perform best.
Context changes where new co-occurrences appear and known disappear (C), again,  is best captured by methods based on significances.
The third data setting (C without usage of $f_{Zipf}$) forced the measurements to handle gap-entries and the inability of the Baseline-method in this concern is revealed.
The produced results include the fact that the context words $w_{51,\ldots,57}$ stay constant in frequency in all time slices.
Under this condition the algorithms Sig, GlobalSig, and MinMax perform best.
Additionally, we tested another case where we set the probability to create a document from $f_1$ 5 times higher than $f_2$.
This causes the number of co-occurrences for $w_{51}, \mathbf{w*}$ to  rise up as well which is caused by the noise introduced by $f_{Zipf}$.
The chance of sampling a word from the noise in combination with $w_{51}$ is also higher when sampling more often from $f_1$ according to formula~\ref{multinom}.
This introduces more co-occurrences with a low significance and the significance-based algorithms tend to diminish the overall volatility (figure~\ref{minmaxsigtriangle}).
For this additional test case we only compared the target functions \textit{triangle} and \textit{sinus}.
Both respective factors use the same amount of maximal changes for the time stamps.
In table \ref{distancetable} we show the mean distances between the volatility values and the expected target function value.
When working with real data (Vocabsize $>>$ 1000) this problem goes to the point, where the resulting context volatility is inversely proportional to the words frequency.
For words with different frequency but similar context changes this results in different values for the context volatility.
The shape of two context volatility time series might be similar but differs in the value range.
When measuring a global volatility among all $T$, 2 words of different frequency classes are not comparable in values even though their contextual change would be comparable.
Such being the case we cannot compare 2 words, which differ in frequency, considering their context volatility in a reliant manner when using significance based algorithms.
%
%
\begin{table}[h]
\centering
\begin{tabular}{l|*{6}{c}r}

          & MinMax & Sig            \\
          \hline
Triangle &0.13&0.27\\          
Sinus    & 0.12 & 0.15  \\
\end{tabular}%
\caption{Mean distances between the calculated volatility and the target function at each point of time; the factor reflecting the triangle function was used 5 times more than the one for sinus}
\label{distancetable}
\end{table}
Comprising the evaluation, the MinMax-algorithm is the most consistent method against differences in term frequency and different types of contextual change. 
Even though we draw our conclusion based just on a synthetic dataset, we do so knowing that the synthetic dataset was specially designed to make the problems which arise when working with real data measurable.  
\section{\uppercase{Example on real data}}
\label{sec:realexample}
\begin{figure}[h]
	\includegraphics[width=\linewidth]{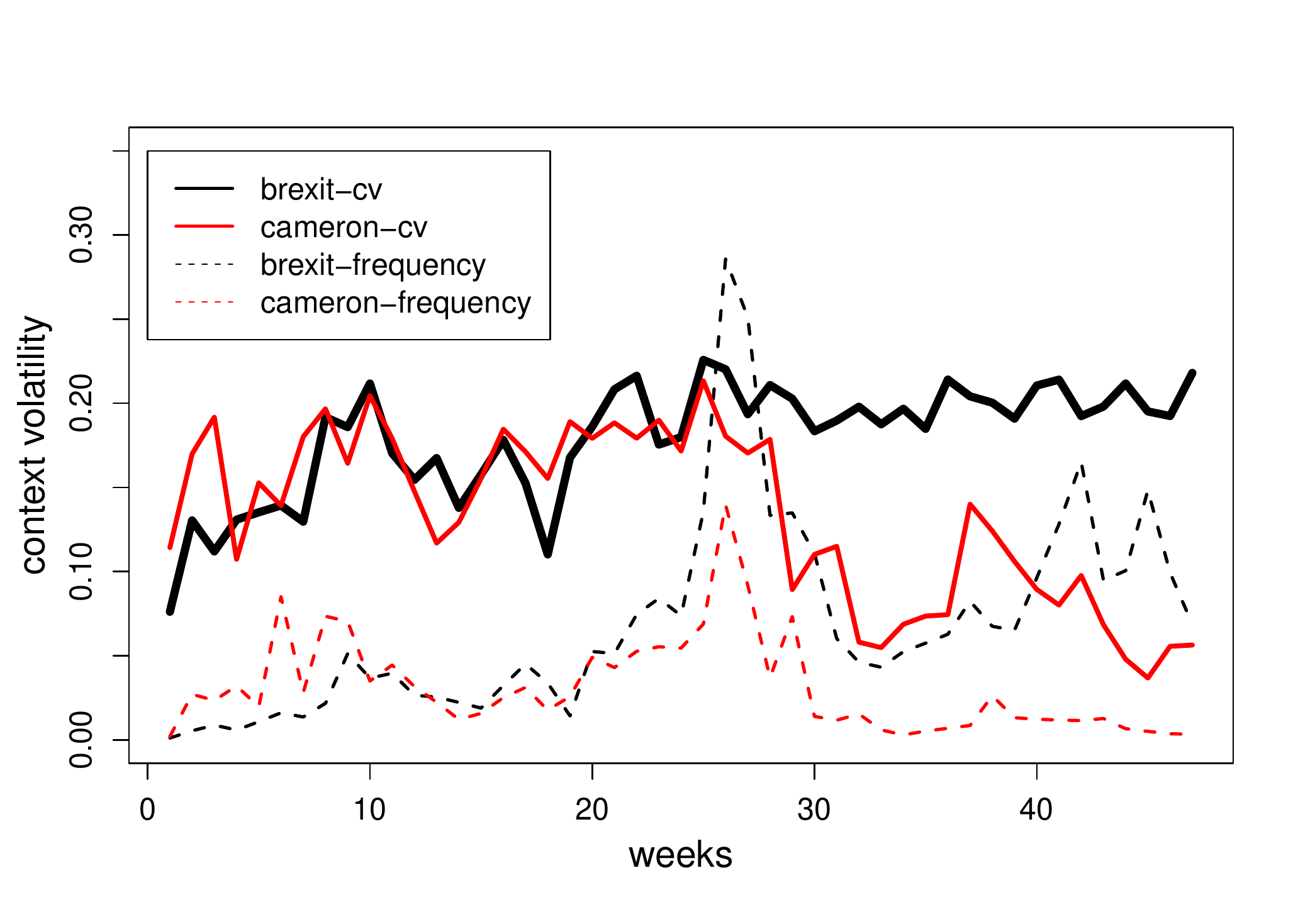}
	\caption{Context volatility for the words \textit{brexit}, \textit{cameron} and their frequencies}
	\label{brexit2}
\end{figure}

In this section we apply the MinMax-algorithm to process 8156 British newspaper articles from January 1st until November 30th of 2016 which include the words \textit{brexit} and \textit{referendum}.\footnote{We used the Guardian API (\url{http://open-platform.theguardian.com/}) to acquire the documents.}
The used language is English, but the concept of context volatility is language independent.
The subject of interest for this analysis of context volatility are keywords surrounding the ``Brexit`` referendum in 2016.
We split the documents into sentences, removed stopwords and used stemming.
Furthermore we split the data set into time slices of weeks.
We used the dice coefficient as significance-weight for the co-occurrences in sentences.
The MinMax-algorithm was applied using a span of 3 weeks ($h=3$).
\begin{figure}[h]
	\includegraphics[width=\linewidth]{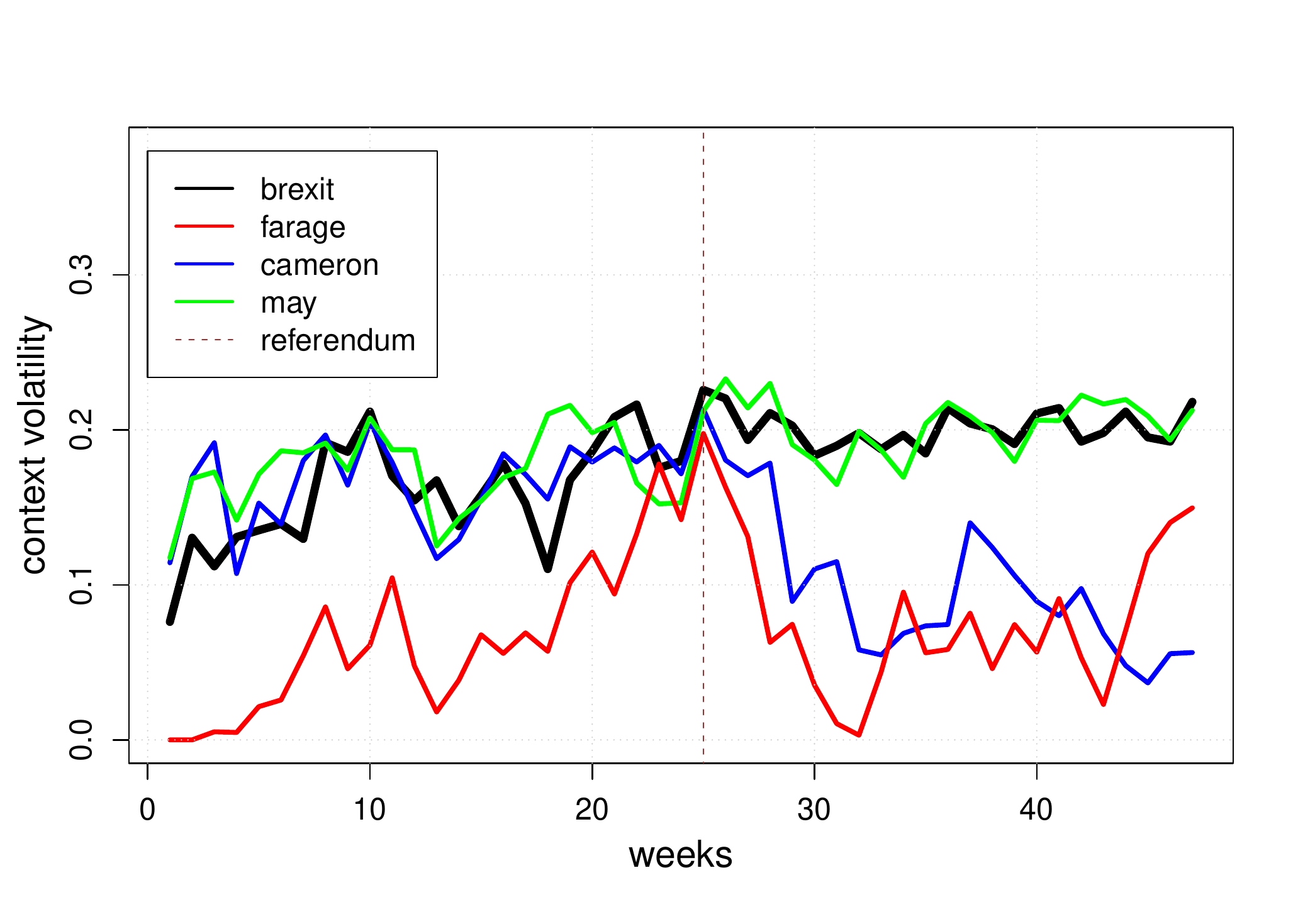}
	\caption{Context volatility for the words \textit{brexit}, \textit{cameron}, \textit{may} and \textit{farage}}
\label{brexit1}
\end{figure}
In figure \ref{brexit1} the measured context volatilities for the words \textit{ brexit, farage, cameron} and \textit{may} are shown.
The time of the referendum is included as dotted slash.
One can identify different ways of contextual change for the words.
For example, the context for the word \textit{cameron} reveals some high amount of change, which suddenly drops right after the referendum.
This goes along with the announcement of Camerons resignation after the referendum. There is basically no new or altered information resulting in a decrease of context volatility.
In this sense, context volatility seems also to track the currentness of information in consecutive time stamps.
In comparison to that we can see  an allover high value for \textit{may}, which is even a little bit higher after the referendum.
\textit{Theresa May}, being the designated prime minister might explain these results.
In figure \ref{brexit2} the volatility values for \textit{brexit} and \textit{cameron} are shown with their respective frequencies.
Although there is some correlation between volatility and frequency, it is obvious that there are lots of phenomena in the volatility values, which can't be explained just by  frequency.
Especially for the word \textit{brexit}, we notice continuously a high volatility value irrespective of its frequency.
In the time immediately around the referendum the frequency of both words have their highest peaks, which is caused by the amount of articles released at that time span.
However, we measure almost the same amount of contextual change for \textit{brexit} already in February just after Cameron announced the referendum, indicating at that stage a high degree of public controversy.
%
%
Using the MinMax-algorithm, we can compare a word's volatility with its frequency over time as well as 2 words that differ in frequency.
\section{\uppercase{Conclusion}}
In this work we presented an evaluation and practical application of the context volatility methodology. 
We've shown a solution for evaluating contextual change measurements, especially the context volatility measure, by creating a synthetic data set, which can simulate various cases of contextual change.
Also, we introduced alternative algorithms, which are superior to the baseline context volatility algorithm, when facing problems on real diachronic corpora (co-occurrence gaps).
In the evaluation and application the MinMax-algorithms shows its robustness when competing with other methods by the ability to use the gap information directly and handle the dynamics in the number of co-occurrences for a word. 
With those improvements in the calculation of context-volatility, we believe that the measure is able to produce new results in various applications.

\vfill
\bibliographystyle{apalike}
{\small
\bibliography{emnlp2017}}

\section*{\uppercase{Appendix}}

\begin{figure}[H]
\includegraphics[width=\linewidth,height=150pt]{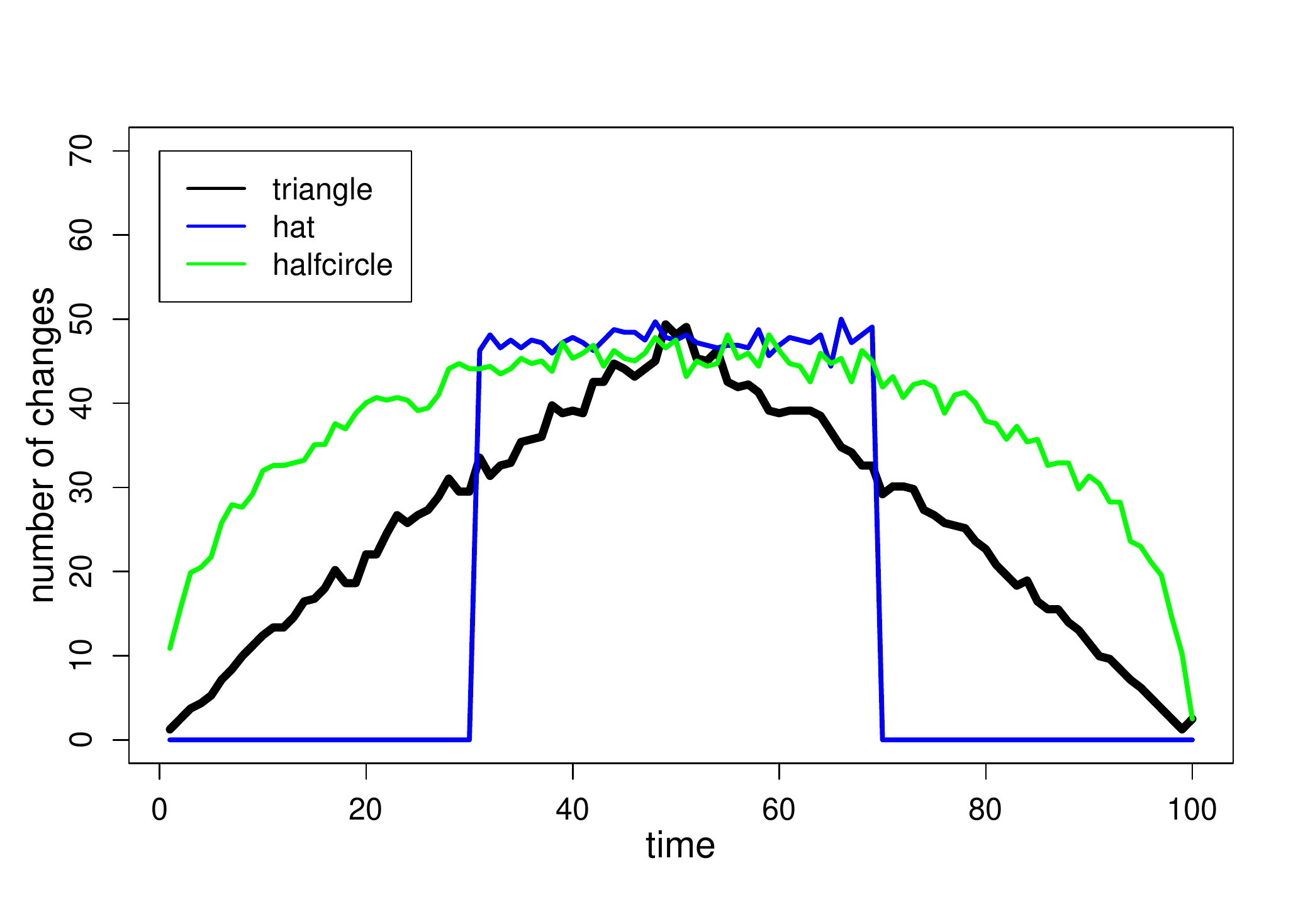}
\caption{Three target functions showing the amount of changes in the function factors in dependence to the time}
\label{targetfunctions_1}
\end{figure}

\begin{figure}[H]
\includegraphics[width=\linewidth]{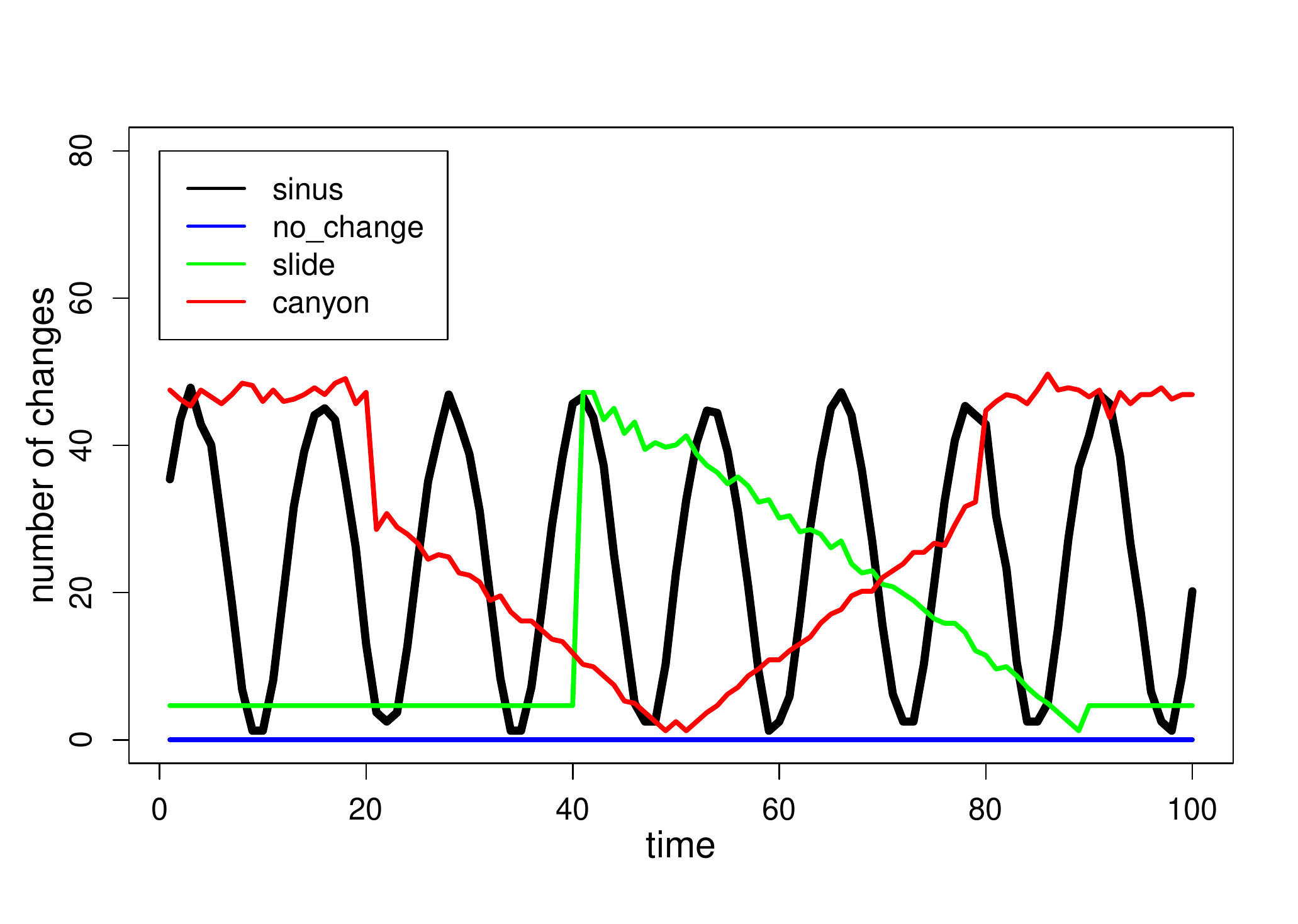}
\caption{Other 4 target functions showing the amount of changes in the function factors in dependence to the time}
\label{targetfunctions_2}
\end{figure}

\vfill
\end{document}